\documentclass[letterpaper, 10 pt, conference]{ieeeconf}  

\IEEEoverridecommandlockouts
\usepackage{times}

\usepackage{soul}
\usepackage{url}
\usepackage[utf8]{inputenc}
\usepackage[small]{caption}
\usepackage{graphicx}
\usepackage{amsmath}
\usepackage{booktabs}
\usepackage[linesnumbered,ruled,vlined]{algorithm2e}
\let\labelindent\relax
\usepackage{enumitem}
\usepackage{caption}
\usepackage{subcaption}

\DeclareMathOperator*{\argmax}{arg\,max}

\title{\LARGE \bf
Joint Mind Modeling for Explanation Generation in Complex Human-Robot Collaborative Tasks
}

\author{Xiaofeng Gao$^{1*}$ Ran Gong$^{1*}$ Yizhou Zhao$^{1}$ Shu Wang$^{1}$ Tianmin Shu$^{2}$ Song-Chun Zhu$^{1}$
	\thanks{$^{1}$ Center for Vision, Cognition, Learning, and Autonomy, UCLA. Emails: \{xfgao, nikepupu, yizhouzhao, shuwang0712\}@ucla.edu, sczhu@stat.ucla.edu.}%
	\thanks{$^{2}$ Massachusetts Institute of Technology. Email: tshu@mit.edu.}
	\thanks{* Equal contributions}
}

\begin{document}

\maketitle
\thispagestyle{empty}
\pagestyle{empty}

\begin{abstract}

Human collaborators can effectively communicate with their partners to finish a common task by inferring each other's mental states (e.g., goals, beliefs, and desires). Such mind-aware communication minimizes the discrepancy among collaborators' mental states, and is crucial to the success in human ad-hoc teaming. We believe that robots collaborating with human users should demonstrate similar pedagogic behavior. Thus, in this paper, we propose a novel explainable AI (XAI) framework for achieving human-like communication in human-robot collaborations, where the robot builds a hierarchical mind model of the human user and generates explanations of its own mind as a form of communications based on its online Bayesian inference of the user's mental state. To evaluate our framework, we conduct a user study on a real-time human-robot cooking task. Experimental results show that the generated explanations of our approach significantly improves the collaboration performance and user perception of the robot. Code and video demos are available on our project website:  \url{https://xfgao.github.io/xCookingWeb/}.

\end{abstract}

\section{Introduction}
In recent years, there has been a great amount of success on building powerful artificial intelligence (AI) systems to solve complex tasks \cite{levine2016end,bansal2017emergent}. As highly autonomous robots are being developed, there is a growing need to make them quickly understood to avoid consequences caused by misunderstanding \cite{gunning2017explainable}. However, existing robot systems are often not human compatible -- i) they do not understand humans' minds and ii) they are just black boxes to humans too. Such limits prevent the AI systems from working with humans effectively. 



\begin{figure}[!t]
	\includegraphics[width=0.49\textwidth]{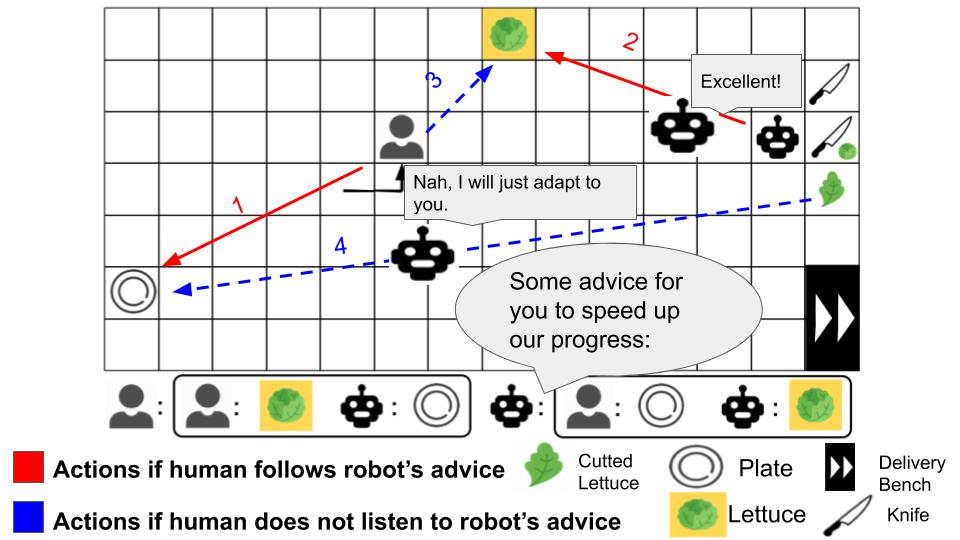}
	\caption{The task \textit{making salad} requires team members to take three lettuce from the basket and cut each one with a knife, before it can be put into the plate and served. After the first lettuce has been cut, the robot is cutting the second one. The robot can identify human's sub-optimal behavior (taking new lettuce from the basket) before generating explanations to the human. }
	\label{fig:teaser}
\end{figure}

Inspired by studies on the Theory-of-Mind \cite{premack1978does,dennett1989intentional}, we believe that a crucial step towards building human compatible systems, particularly for human-robot collaborations, is to understand human activities and their underlying mental state.  As a motivating example, consider a robot chef helping a human make salads in the kitchen shown in Figure \ref{fig:teaser}. Even when the robot understands how to perform the task on its own, it would be challenging to finish the task efficiently without having a shared mental model with its human partner. For making the salad, the robot believes the plate should be picked up by the user while the human agent believes the other way. If the robot can identify such discrepancies between different agents' mental states, it can generate explanations to mitigate the differences and encourage the correction of sub-optimal human behavior.

To this end, we propose a framework that improves human-robot teaming performance through explanations. With a graph-based representation, the robot can maintain the mental states of both team members during a highly-structured collaborative task. The robot can then generate explanations when difference between mental states is detected, which implies sub-optimal user behaviors. In summary, the main contribution of this paper is three-fold:
\begin{itemize}[leftmargin=*]
	\item We design a real-time collaborative cooking game as an online user study system and develop an evaluation protocol, which can be accessed from our website.
	\item We propose to understand complex human activities using an action parsing algorithm based on an And-Or graph task representation, which allows the robot to infer human mental states in complex environments.
	\item Based on the inferred human mental state, we propose an explanation generation framework. Experiments on a real-time cooking task show that our approach successfully improves user perception of the robot and leads to better human-robot collaborations.
\end{itemize}

\section{Related Work}
\noindent \textbf{Human-aware planning.} 
Designing robots that can work with humans has been widely studied by researchers. Most of the prior works hope to create robots to better understand and adapt to human collaborators. \cite{liu2016goal} evaluates a collaborative task allocation framework based on a Bayesian inference of human intention. \cite{hadfield2016cooperative} proposes a formulation of the value alignment problem assuming the robot learning an unknown human reward function. Optimal solutions can be achieved when the human demonstrates active teaching behavior. To deal with sensor uncertainty and task ambiguity in a collaborative assembly task, \cite{hawkins2014anticipating} uses an And-Or tree structure as the task representation, which is similar to our approach. When sub-optimal user behavior are encountered, \cite{reddy2018you} proposes to learn the incorrect human internal dynamics model via inverse RL and then perform an internal-to-real dynamics transfer to assist users in shared-autonomy tasks. Our framework differs from this line of research in that we also aim at improving humans' understanding of robots' models using communicative actions. Such two-way understanding will further help human-robot collaborations.

\noindent \textbf{Goal-driven explainable AI.} 
In contrast to data-driven XAI which improves understanding of "black-box" machine learning algorithms given input data, goal-directed XAI typically explains the behavior of an agent or robot for a specific task \cite{langley2017explainable,anjomshoae2019explainable,miller2019explanation}, in order to increase model transparency \cite{struckmeier2019autonomous}, human's trust \cite{wang2016trust} or task performance \cite{xu2015optimo}. Some of the works achieve this aim by enabling robots to directly generate easy-to-understand motions \cite{dragan2013generating,kwon2018expressing} or task plans \cite{zhang2017plan}. Other works, similar to ours, focus on using explicit communication to change user mental state, e.g., updating users' incorrect reward functions \cite{tabrez2019explanation}, correcting users' false belief or misunderstanding about the environment \cite{gong2018behavior, sreedharan2018hierarchical}, resolving the disagreement between collaborators' actions \cite{nikolaidis2018planning} or providing users with necessary knowledge about the current situation \cite{devin2016implemented}. Compared to these work that often require offline training with humans or theoretical assumptions on the human models, this paper takes a direct approach to generate explanations solely based on an online estimation of human model and knowledge of the task structure. The experiment results show our approach is empirically effective in an ad-hoc human-robot teaming settings \cite{stone2010ad} where pre-coordination is not available.  

\begin{figure}[t]
	\includegraphics[width=0.5\textwidth]{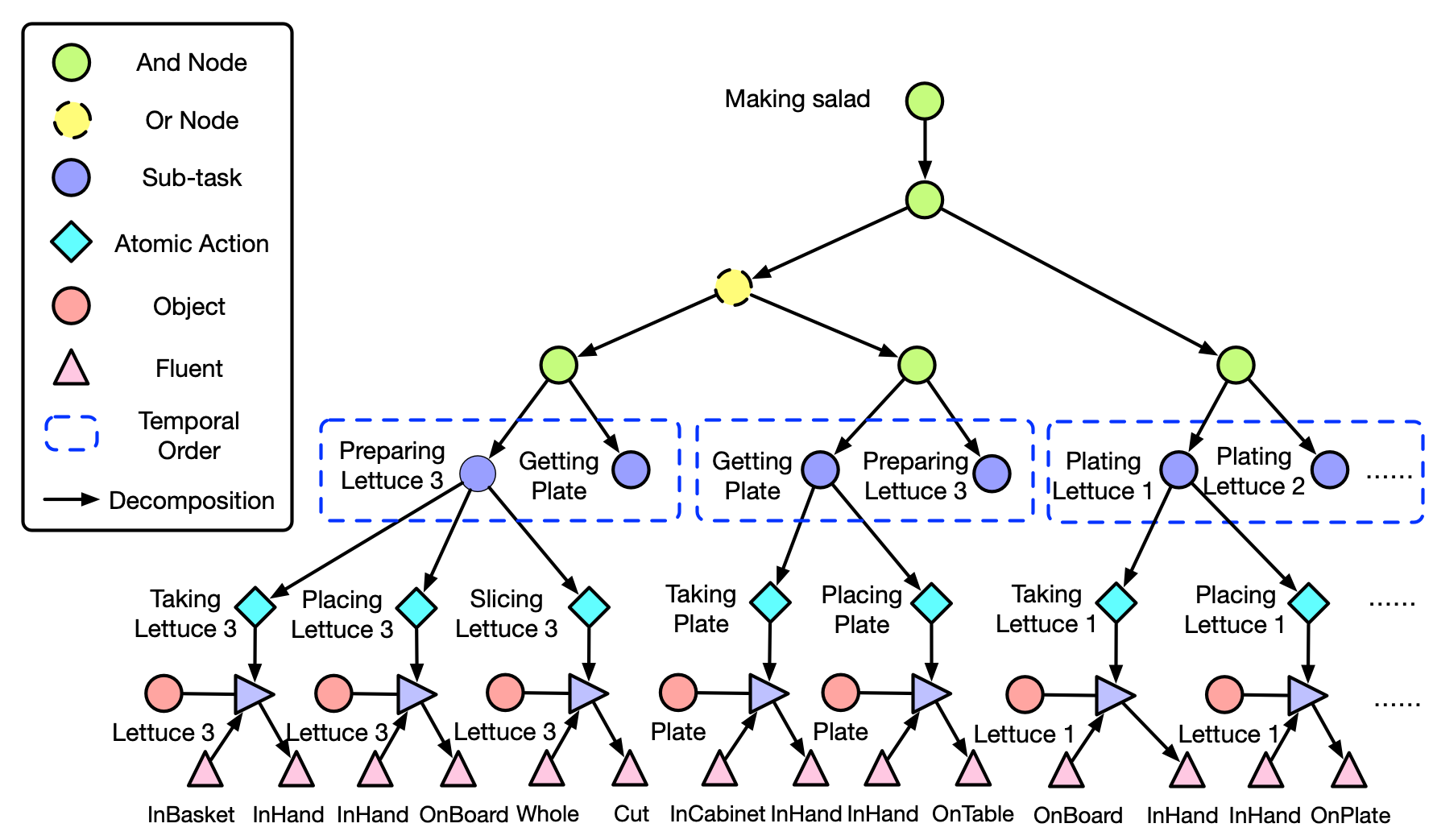}
	\caption{The hierarchical mind model for the collaboration task, "making salad", represented by an AoG. The And node represents temporal relations between sub-tasks. The Or node represents two possible ways for the team to finish the tasks. Each terminal node (diamond) denotes an atomic action that would cause certain fluent changes (triangles) for objects.}
	\label{fig:teaser_aog}
	\vspace{-2pt}
\end{figure}

\begin{figure*}[t]	
	\includegraphics[width=0.9\textwidth]{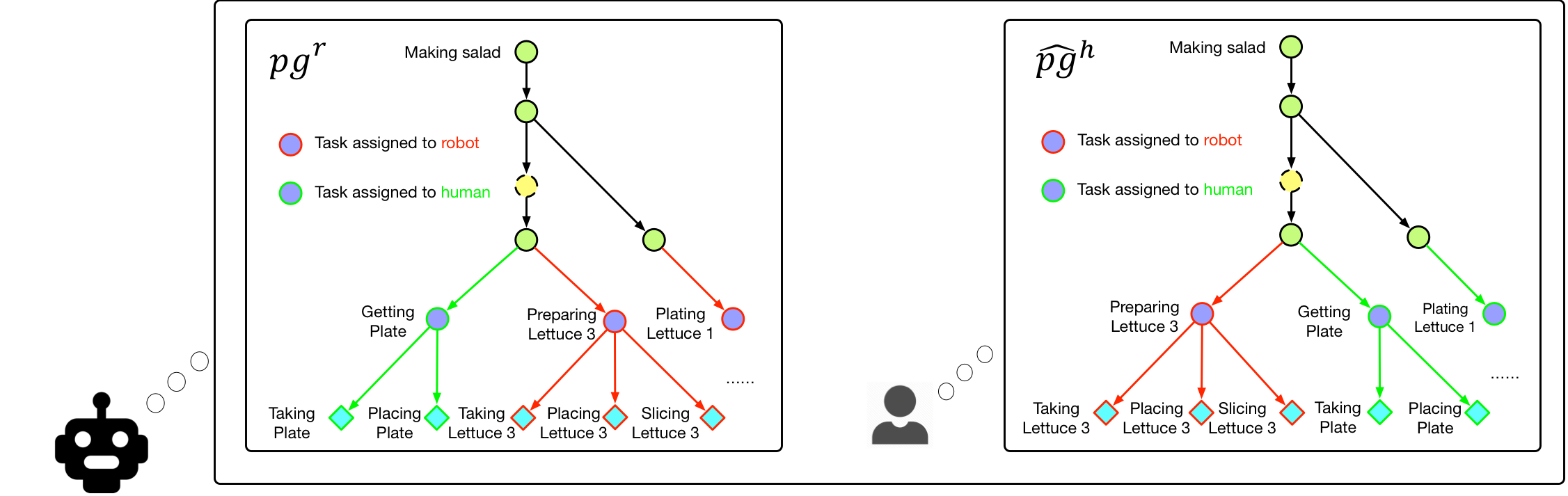}
	\caption{Robot mental state $pg^{r}$ and inferred human mental state $\hat{pg}^{h}$ represented as parse graphs. }
	\label{fig:teaser_pg_new}
\end{figure*}

\section{Single Agent Mind Model}
And-Or graphs (AoGs) have been widely used for robot task planning \cite{xiong2016robot,shu2017learning,liu2018interactive} and human activity modeling \cite{tu2013unsupervised,shu2015joint}. As a hierarchical representation, a spatial-temporal-causal And-Or graph (STC-AoG) encodes a joint task plan and corresponding spatial, temporal, and causal relations an agent could have about the task \cite{xiong2016robot}. In this work, we propose to use a STC-AoG as a unified representation of a robot's knowledge and plan regarding the task as well as the inferred human's knowledge and plan. An example of a single-agent plan for \textit{making salad} is in Figure \ref{fig:teaser_aog}.

\subsection{STC-AoG as a Hierarchical Mind Model}
In general, an And-Or Graph consists of nodes and edges. The set of nodes includes Or node, And node, and Terminal node. Each \textbf{Or node} specifies the Or relation: only one of its children nodes would be performed at a given time. An \textbf{And node} represents the And relation and is composed of several children nodes. Each \textbf{Terminal node} represents a set of entities that cannot be further decomposed. The edge represents the top-down sampling process from a parent node to its children nodes. The root node of the And-Or tree is always an And node connected to a set of And/Or nodes. Each And-node represents a sub-task which can be further decomposed into a series of sub-tasks or atomic actions.

In this paper, the graph $G=<A,F,T,V,R,P>$ is formally defined as the following:
\begin{itemize}[leftmargin=*]
	\item $A$ is a set of terminal nodes. Each node corresponds to an atomic action $a \in A$.
	\item $F$ is a set of object states essential to the task, including possible pre-conditions and post-effects of atomic actions.
	\item $T: F \times A \rightarrow F$ is a set of transition rules that represent state changes caused by atomic actions.
	\item $V$ is a set of non-terminal nodes, which can be further decomposed into two sets: the And nodes $S$ and the Or nodes $O$. Each sub-task corresponds to an And node $s$, which encodes a temporal relationship between its children. An Or node $o$ forms a production rule with an associated probability, i.e. you may choose one of its children each weighted with a certain probability. 
	\item $R$ is the set of production rules.
	\item $P$ is the set of probabilities on production rules.
\end{itemize}

\noindent \textbf{Causal relation.} 
Causal knowledge represents the pre-conditions and the post-effects of atomic actions. We define it as a fluent change caused by an action. Fluent $f \in F$ can be viewed as some essential properties in a state $x$ that can change over time, e.g., the temperature in a room and the status of a heater. For each atomic action, there are pre-conditions characterized by certain fluents of the states. E.g., an agent cannot successfully turn on the heater unless it is plugged in. As the effect of an action, certain fluents would be changed, and the state $x$ would evolve to $x'$. For example, if someone turns on a heater, the temperature of the room will be higher (and the heater would be on). It is formulated as one of the transition rules $T$. 

\noindent \textbf{Temporal relation.}
Temporal knowledge encodes the schedule for an agent to finish each sub-task. It also contains the temporal relations between atomic actions in a low level sub-task. The sub-task preparing salad, for example, consists of taking salad, placing it onto the cutting board, and using the knife.

\noindent \textbf{Spatial relation.}
Spatial knowledge represents the physical configuration of the environment that is necessary to finish the task. In our case, to make the salad, an agent needs to know the locations of ingredients (e.g., lettuce), tool benches (e.g., basket, cutting board), delivery benches, etc.

\subsection{Parse Graphs as Mental State Representations}

During the collaboration, an agent can use parse graphs to represent the mental states of itself or the other agent. A parse graph is an instance of an And-Or Graph, each of its Or nodes selects one child node. Figure \ref{fig:teaser_pg_new} shows two parse graphs represent the robot and human's plan for the situation shown in \ref{fig:teaser}. In our case, the parse graph ${pg}_{t}=<s^{h}_{t}, s^{r}_{t}, a^{h}_t, a^{r}_t, f^{h}_{t}, f^{r}_{t}>$ is one possible plan for both agents to finish the task. Particularly, the root node leads to a selection of individual sub-tasks $(s^{h}_{t}, s^{r}_{t})$ as sub-goals assigned to human and robot agent. To achieve these sub-goals, agents perform atomic action $(a^{h}_t, a^{r}_t)$ based on their belief of current fluent $(f^{h}_{t}, f^{r}_{t})$. 

\subsection{Joint task planning by parsing STC-AoG}
\label{planning}
To construct the mental state representation for the robot, we design an algorithm based on STC-AoG parsing to select the optimal task plan for the team.

Given a set of sub-tasks $S$ necessary to complete the joint task, the objective is to minimize the total task completion time by assigning a sub-task to either a robot or human agent, without violating any latent constraint:
\begin{equation}
\begin{aligned}
	\min_{x_{s}^{v}, \tau_{s}} \max_{v \in \{r, h\}} \sum_{s \in S} x_{s}^{v} \delta_{s}^{v} \\
	\text{s.t.}~x_{s}^{v} \in X_\text{feasible}, \tau_{s} \in {\Gamma}_\text{feasible}.
\end{aligned}
\end{equation}
where $x_{s}^{v}$ is a binary variable indicating whether to assign sub-task $s$ to agent $v$, and $\tau_{s}$ is a continuous variable representing the finishing time for the sub-task $s$. Constant $\delta_{s}^{v}$ represents the amount of time for agent $v$ to finish the sub-task $s$. $X_\text{feasible}$ and $\Gamma_\text{feasible}$ represent the set of valid assignments that satisfies latent causal constraints, e.g., an agent cannot hold two objects at the same time; a sub-task can be performed only if pre-conditions are met; after all assigned sub-tasks have been completed, the final state should satisfy the goal requirement.

We search for the optimal task plan via a dynamic programming algorithm. Starting from the initial state $f_{b}$, we make valid sub-tasks assignments and simulate new intermediate state $f_{e}$ based on the state transition function $T$. By updating the current optimal consumed time and the corresponding sub-task assignment vectors for every intermediate state, our algorithm will finally reach the optimal plan for the entire task. During the updating process, we also record the sub-task assignment vectors for previous states, in order to generate the whole optimal assignment $\{x^{v}_{s}\}_{s=1, ..., |S|}$ and completion time $\tau_{1}, ..., \tau_{|S|}$ for each sub-task. 
After the task plan is computed, the robot's mental model is represented by a parse graph, as shown in the left part of \ref{fig:teaser_pg_new}: each sub-task in the task plan indicates a sub-goal that an agent needs to achieve at the time being. Sub-tasks are further connected with a sequence of corresponding atomic actions, which have certain pre-conditions and post-effects.

\section{Joint Mind Modeling for Human-Robot Collaborations}
Our goal is to enable efficient human-aware collaboration for a human-robot team. Specifically, robots need to understand human agents based on their actions and decide whether the team is moving in the right direction. We propose to model the robot mental state $pg^{r}$ and the human mental state $pg^{h}$.

\subsection{Mind Models for Human and Robot}
We treat the robot's mind as the oracle, i.e., it contains all necessary spatial, temporal, and causal information the team needs to finish the task. For example, at any given time $t$, the robot has a certain expectation of (i) current low level sub-goals $(s^{h}_{t}, s^{r}_t)$ both agents should be pursuing; (ii) the actions $(a^{h}_{t}, a^{r}_t)$ agents should perform; (iii) whether current object fluents satisfy pre-conditions of such actions, and what would be the post-effects. 

It is also necessary to model the user's mind, which acts as a strong inductive bias in predicting user activities. As the user's mental state $pg^{h}_{t}$ is not directly available to the robot, we propose to infer it from user behavior and the history of communication.

\begin{figure}[t]
    \centering
	\includegraphics[width=.4\textwidth]{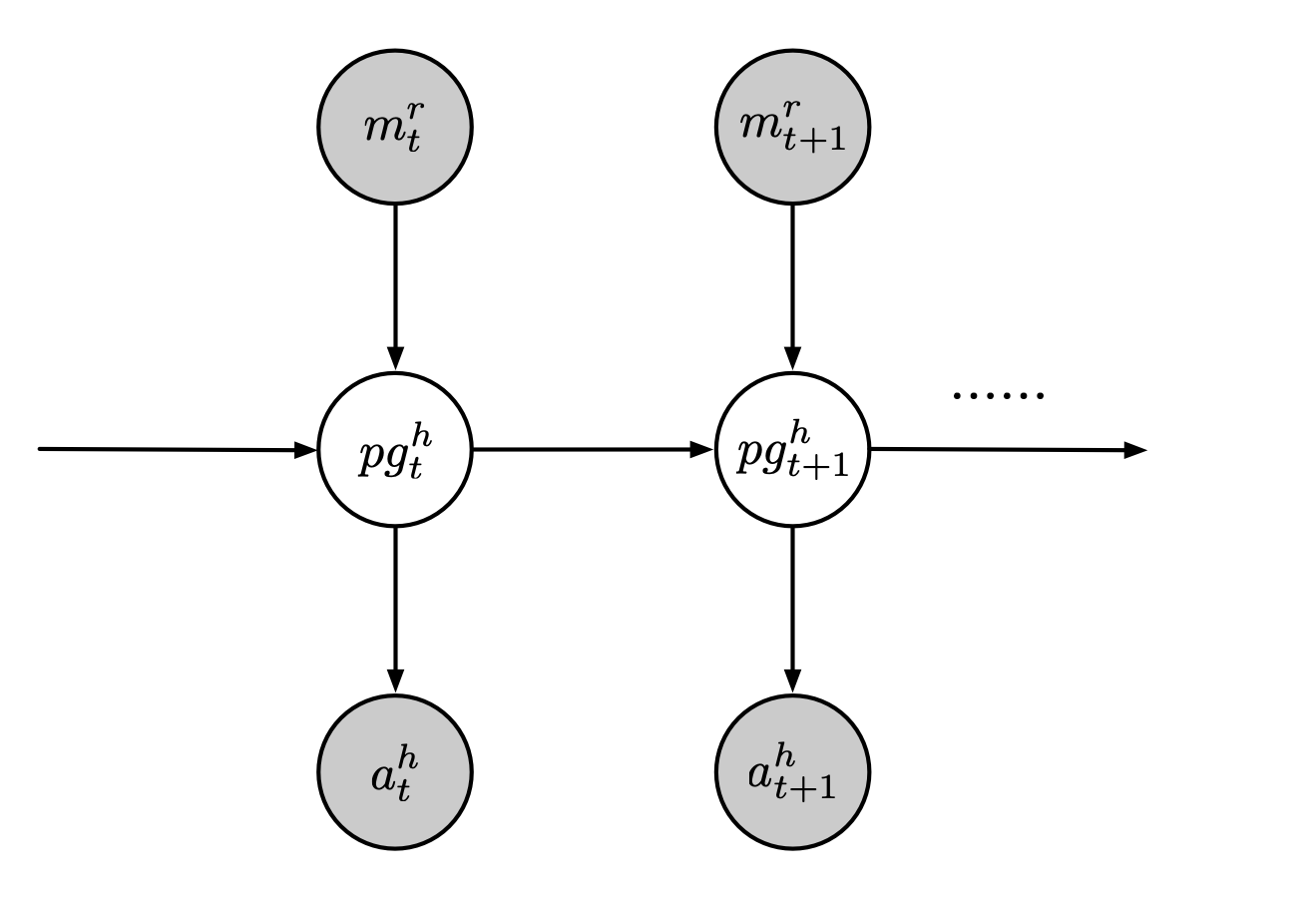}
	\caption{Human mental model update process. We use it to infer user mental state ${pg}^{h}$, which is hidden to the robot. Here we assume human actions $a^{h}_{t}$ and robot message $m^{r}_{t}$ are conditional independent given human mental state ${pg}^{h}_{t}$ at time $t$. }
	\label{fig:graph_model}
	\vspace{-3pt}
\end{figure}

\subsection{Human Mental State Inference}
\label{sect:h_model}

\begin{figure*}[t]
	\centering
	\includegraphics[width=0.9\linewidth]{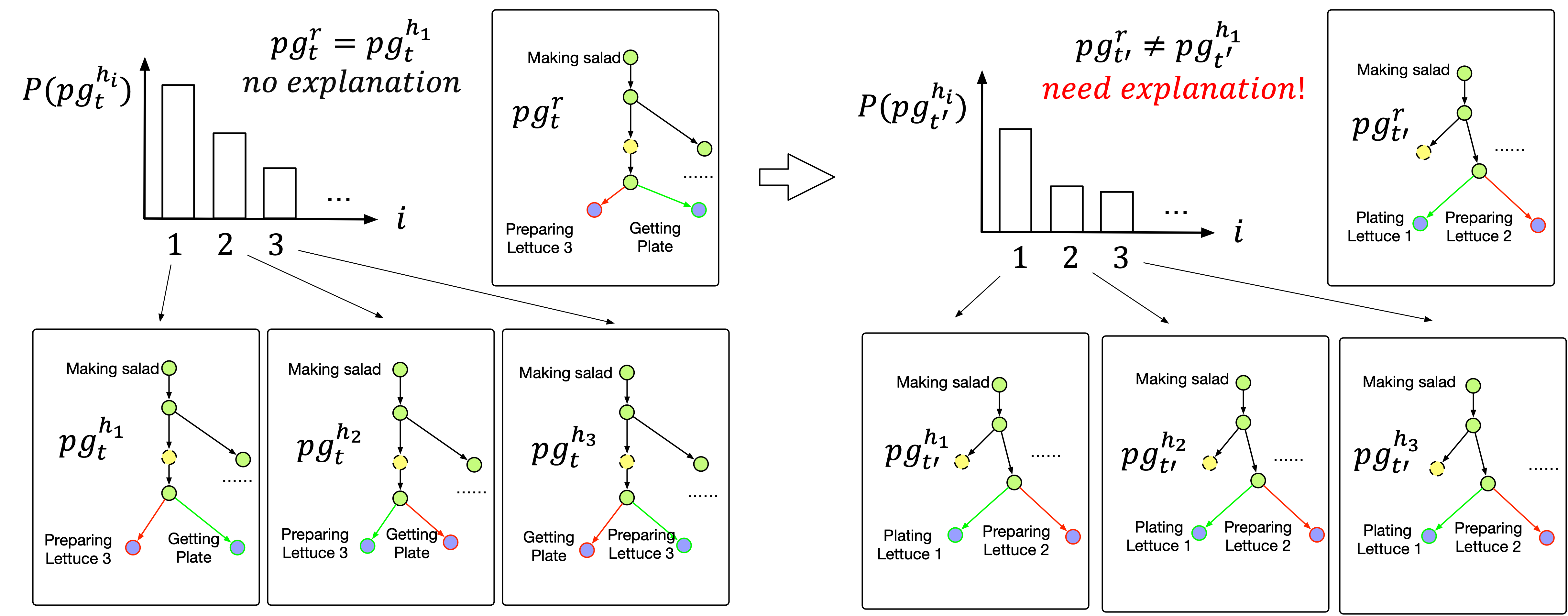}
	\caption{Explanation timing. At time $t$, sort posterior probability of $pg^{h_{i}}_{t}$ in descending order, and then compare the most possible user mental state $pg^{h_{1}}_{t}$with robot mental state $pg^{r}_{t}$. Since they are the same, there is no need to explain to the user. At time $t'$, $pg^{h_{1}}_{t'}$ is not equal to $pg^{r}_{t'}$, therefore, the robot should provide the explanation.
	}
	\label{fig:explanation_timing}
	\vspace{-5pt}
\end{figure*}

Based on the observed user behavior, we infer the most likely human mental state $\hat{pg}^{h}$, including the belief, goal and action plans. On a high level, this inference process uses observed user actions and communication history to infer human mental state. Specifically, given the And-Or graph $G$ and human-robot interaction data $D_{T} = \{ d_{t} \}_{t=1,...,T}$, we infer the user mind $\hat{pg}^{h}$ iteratively:
\begin{align}
	\hat{pg}^{h} &= \argmax_{{pg}^{h}} p({pg}^{h}|D_{T}, G),
\end{align}

\vspace{-15pt}

\begin{align}
    p({pg}^{h}|D_{T}, G) &\propto p({pg}^{h}|G, D_{T-1}) p(d_{T}|{pg}^{h}, G).
\end{align}

Here the first term models the prior on the user mind given previous data $D_{T-1}$ and AoG structure $G$. The second term models the likelihood for new data $d_{T}$.

To model the likelihood function $p(d_{T}|{pg}^{h},G)$, we take a sampling-based approach. For each interaction data $d$, we consider user atomic action $a^{h}_{obs}$ and communication between the two agents $m$.  The idea is to model how likely the user performs action $a^{h}_{obs}$ when receiving message from the robot $m^{r}$, with current mental state ${pg}^{h}$, as shown in Figure \ref{fig:graph_model}. Assuming $a^{h}_{obs}$ and $m^{r}$ are conditional independent given ${pg}^{h}$ we have:
\begin{align}
	p(d|{pg}^{h}, G) = p(a^{h}_{obs}|{pg}^{h}, G) p(m^{r}|{pg}^{h}, G), 
\end{align}
\begin{align}
    p(a^{h}_{obs}|{pg}^{h}, G) = \sum_{a^{h}_{samp}} p(a^{h}_{samp}|{pg}^{h}) p(a^h_{obs}|a^{h}_{samp}),
\end{align}
where $p(a^{h}_{samp}|{pg}^{h})$ denotes the probability of sampled human action $a_{samp}^{h}$ given current estimation of human mental state ${pg}^{h}$. $p(a^h_{obs}|a^{h}_{samp})$ measures the similarity between observed human trajectory $a^{h}_{obs}$ and sampled trajectory $a^{h}_{samp}$. 

In practice, we use rapid-exploring random tree (RRT*) for trajectory sampling and dynamic time warping (DTW) based approach to compare trajectories. DTW outputs a difference score $diff$. We use it in the energy function for the Boltzmann distribution. Then we update the human mental state in every time-step through the following equation:
\begin{equation}
	P(\hat{pg}^{h}_{t+1} | D_{T}, G) = \frac{1}{Z}  e^{-\frac{diff}{T}} \lambda^{n } P(\hat{pg}^{h}_{t} | D_{T-1},  G) ,
	\label{equ:update}
\end{equation}
where $T$ is a constant temperature term, $Z$ is a normalization constant, and $\lambda$ ($>1$) is a constant that controls the importance of an explanation. It models how much information the user can retain for an explanation. $n$ is the number of times an explanation about $\hat{pg}^{h}$ is generated for the user in this task. Therefore, $\lambda^{n}$ implicitly encodes the communication history $m$. Right now, we only consider communications from robot to human $m^{r}$. Communication from human to robot $m^{h}$ can be considered in the future by adding corresponding energy terms.  For now, some parameters ($T$ and $\lambda$) are set heuristically. These parameters can be learned from annotated user data \cite{carreira2005contrastive}. 
\begin{algorithm}[t]
	\caption{Planning and explanation generation}
	\label{algo:update}
	\While { Task not finished } 
	{
		
		\If {Replan needed}
		{
			Collect state information from the game\;
			Collect predicted human intentions from the last time step \;
			Call DP planner \;
			Obtain a new sequence of sub-tasks from planner and re-organize AoG based on it\;
			Parse AoG through checking pre-conditions and post-effects against the current environment state information \;
			Find out the next atomic action to execute based on parsing result \;
		}
		Predict human intentions by equation (\ref{equ:update}) \;
		Measure the difference between predicted intention and expected human actions\;
		Generate an explanation if the difference  $>\tau$ \;
		
	}
	
\end{algorithm}

\subsection{Robot Mental State Update}

Based on the observations in the environment, the robot can update its joint task plan. It is a two-step process. First, the robot collects all relevant information about the task and calls a DP planner described in Section \ref{planning} to obtain an optimal sequence of sub-tasks. Then the robot updates its mental state through re-organizing AoG (Delete finished nodes. Re-order unfinished nodes. If necessary, add back nodes deleted previously). Second, the robot uses causal knowledge (pre-conditions and post-effects of each atomic action) in the AoG terminal nodes to determine the next atomic action. If pre-conditions for the next atomic action are satisfied, the robot will execute it. Otherwise, the robot will be idle, waiting for the user to complete the other part of the job.

\section{Explanation-based task coaching}
In this section, we propose a framework for explanation generation to enable efficient human-robot collaboration. 

\subsection{Explanation framework}
As shown in Algorithm \ref{algo:update}, the framework includes an iterative process of online planning and explanation generation: 
\begin{enumerate}[leftmargin=*]
	\item At a given time, the robot updates its mental state to represent the expected current goals of both agents and corresponding atomic actions;
	\item The mental state of the human agent can be inferred, which would be further compared to the robot's mental state. Based on the result, the robot would decide whether explanations are necessary;
	\item On the occasions where users perform an action other than that indicated in the explanation, the robot would update its task plan and mental model to reflect the best joint policy and expected mental models in the new state. 
    
    Take the task \textit{making salad} for example. At the beginning of the game, an optimal plan requires the user to first take the plate. A sub-optimal plan could be the user first taking the lettuce. If the user insists on taking the lettuce first regardless of whether explanations are given, the robot will update the task plan and expect the user to gather the plate afterwards. 

\end{enumerate}

\begin{figure*}[ht]
	\begin{subfigure}[t]{0.47\textwidth}
		\centerline{\includegraphics[width=\textwidth]{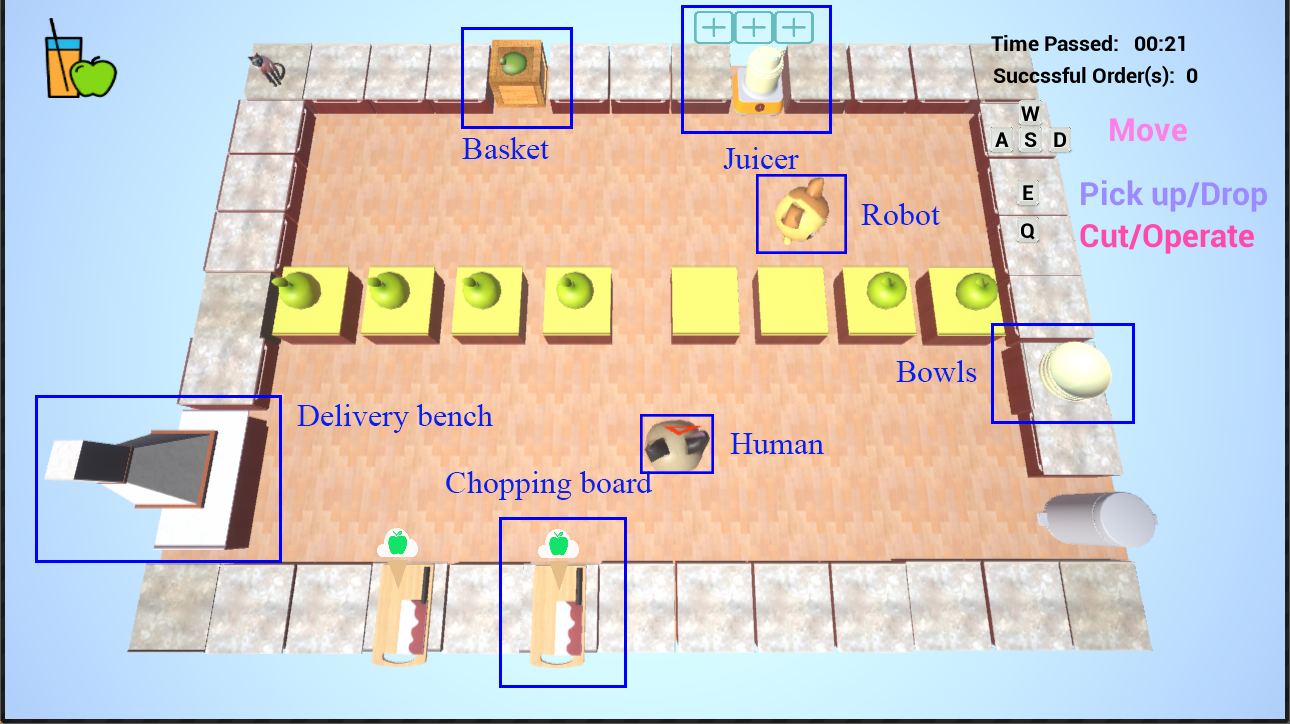}}
		\caption{}
		\label{fig:level}
	\end{subfigure}
	\quad
	\begin{subfigure}[t]{0.47\textwidth}
		\centerline{\includegraphics[width=\textwidth]{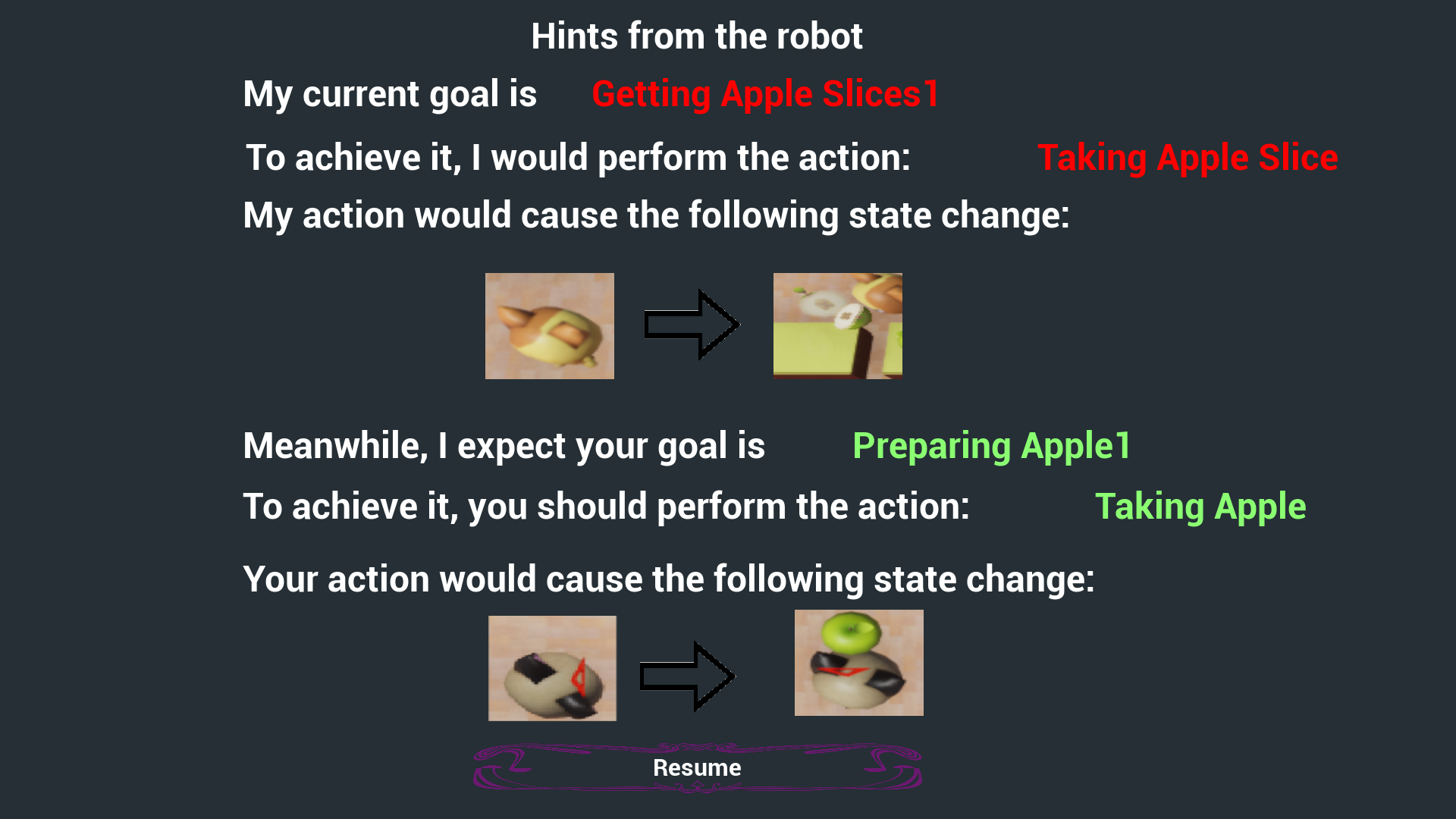}}
		\caption{}
		\label{fig:explanation}
	\end{subfigure}
	\caption{(a) A top-down view of our collaborative cooking game, where the user (the bottom character) collaborates with a robot (the top character) on some cooking tasks, e.g. {\em making apple juice}. (b) The explanation interface exhibits the expected sub-tasks for both agents. Pre-conditions and post-effects of atomic actions are displayed as well.}
	\label{fig:interface}
	\vspace{-5pt}
\end{figure*}
\subsection{Explanation Timing}
The explanation serves to provide users with the knowledge necessary to finish the task efficiently. This is achieved by inferring the user's mental model during the interaction and comparing it with the robot's. Whenever a disparity between these two models is detected, we can generate explanations to encourage correction of the user's mental state.

During collaboration, we use temporal parsing to get robot mental state $pg^{r}_{t}$ from its And-Or graph at time $t$. As in Section \ref{sect:h_model}, user mental states $\hat{pg}^{h}_{t}$ can be inferred based on communication history and action sequences. The system generates explanations when there is a mismatch between the robot mental state and inferred human mental state: $|pg^{r}_{t} - \hat{pg}^{h}_{t}| > \epsilon$.
In practice, we measure $P( \hat{pg}^{h}_{t} | D_{T}, G)$ for every sub-tasks at each time step based on equation (\ref{equ:update}). If the probability $P( \hat{pg}^{h}_{t} = pg^{r}_{t} | D_{T}, G)$ is lower than a threshold $\tau$, we generate an explanation for the user. This process is shown in Figure \ref{fig:explanation_timing}. 

\subsection{Explanation Content}

We envision the disparity occurred between the user's mental state and robot's due to several reasons:

\begin{enumerate}[leftmargin=*]
	\item The user wants to achieve goals that are different from the robot's expectation;
	\item The user performs incorrect atomic actions to achieve a sub-goal;
	\item The user is unaware of the pre-condition or effect of an atomic action.
\end{enumerate}

In this paper, we do not distinguish between the possible causes of disparity when choosing the explanation timing, as they are too ambiguous. Instead, we propose to generate hierarchical explanation which consists of three components of the robot's mind representation:
\begin{enumerate}[leftmargin=*]
	\item The robot would explain the current expected sub-goals of both agents $(s^{h}_{t}, s^{r}_{t})$ based on its mental state $pg^{r}$, e.g., "My current goal is preparing the lettuce. Meanwhile, your expected goal is getting the plate."; 
	\item The robot communicates the expected atomic actions that both agents are supposed to perform $(a^{h}_{t}, a^{r}_{t})$, e.g., "Currently, I'm performing the action slicing the lettuce. You are supposed to perform the action taking the plate."; 
	\item In addition, by showing images of world states before and after an action (as shown in Figure \ref{fig:explanation}), the robot would also demonstrate the fluent change caused by an atomic action $f_{t} \xrightarrow[]{a_{t}} f_{t+1}$. 
\end{enumerate}

\begin{figure*}[ht]
	\centering
	\includegraphics[width=0.9\linewidth]{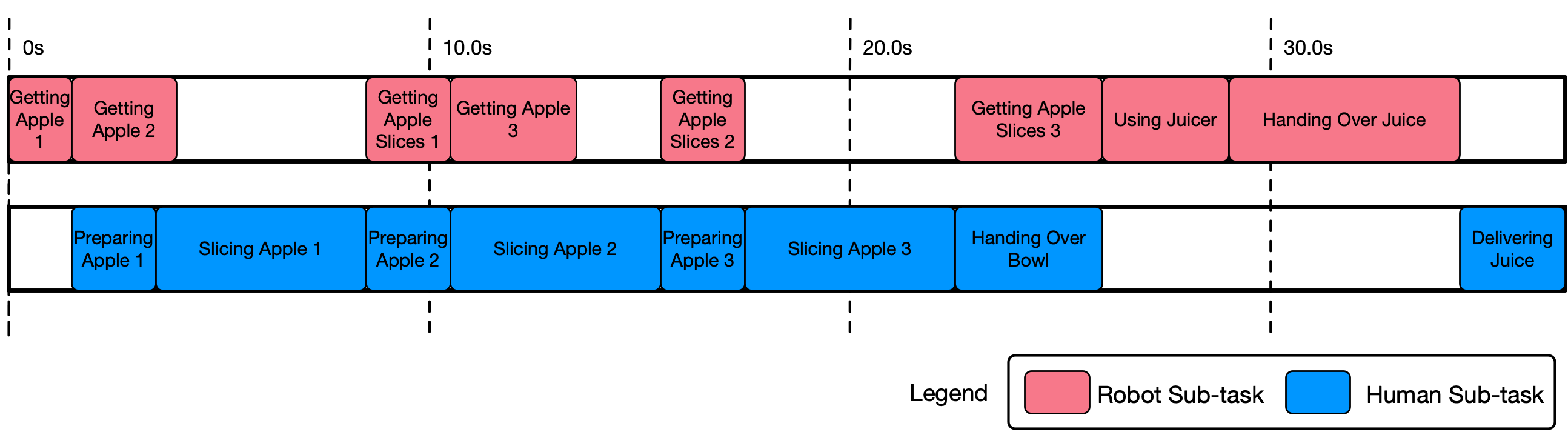}
	\caption{An example task schedule for {\em making apple juice}. The robot maintains the schedule to reflect its expectation on how the team should finish the task. Each color block represents a sub-task, performed by either robot or human. At a specific timing, we can assign tasks to both agents based on the schedule. E.g. at 10.0s, the robot is {\em getting apple slices 1} while the user is supposed to be {\em preparing apple 2}. The schedule gets updated based on inferred human mental states, as shown in Algorithm \ref{algo:update}.}
	\label{fig:schedule}
	\vspace{-5pt}
\end{figure*}

\section{User Study}
We conducted a user study in a gaming environment to evaluate our algorithm, where participants can collaborate with agents on a virtual cooking task. The gaming environment and explanation interface are displayed in Figure \ref{fig:interface}.

\subsection{Experiment Domain}
Our experiment domain is inspired by the video game \textbf{Overcooked}\footnote{\url{http://www.ghosttowngames.com/overcooked/}}, where multiple agents are supposed to make use of various tools and take different roles to prepare, cook, and serve various dishes. Particularly, we use Unreal Engine 4 (UE4) to create a real-time cooking task, namely making \textit{apple juice}. To finish the task, teammates need to take apples from the box and slice them with a knife near the chopping board. Three apple slices should be put into the juicer before producing and delivering apple juice. Figure \ref{fig:level} shows a top-down view of the environment. The game interface is designed to be interactive (e.g., object appearance will change after taking valid actions) so that people can easily play through.

To finish the task, each user needs to complete a sequence of 62 atomic actions, if acting optimally, and observe 5 different object fluent changes with a total state space around $10^9$. An example task schedule is shown in Figure \ref{fig:schedule}.

\subsection{Experiment Design}
\textbf{Hypotheses.} 
The user study tests the following hypotheses with respect to our algorithm in the collaboration:
\begin{itemize}[leftmargin=*]
	\item \textbf{H1: Task completion time.} Participants would collaborate with the robot more efficiently if the robot generates explanations based on the human mental state modeling, compared to the other conditions.
	\item \textbf{H2: Perception of the robot.} Participants would have higher perceived helpfulness and efficiency of the robot, as a result of receiving explanations based on the human mental state modeling, compared to the other conditions.
\end{itemize}

\noindent \textbf{Manipulated Variables.} 
We use a between-subject design for our experiment. In particular, users are randomly assigned to one of three groups and receive different explanations from the robot:

\begin{itemize}[leftmargin=*]
	\item \textbf{Control:} Users would not get any explanations from the robot. As a result, they can learn to finish the task by interacting with the environment.
	\item \textbf{Heuristics:} The robot gives explanations when there is no detected user action for a period of time. This serves as a simple heuristic for the robot to infer whether the user is having difficulties in finishing the task. The timing threshold is set to 9.3 seconds, based on the result of a pre-study in which users can actively ask for explanations when they get stuck.
	\item \textbf{Mind modeling:} The robot gives explanations when there is a disparity between robot and  human mental states.
\end{itemize}

\noindent \textbf{Study Protocol.}
Before starting the experiment, each participant signs an informed consent form. An introduction is given afterward, including rules and basic controls of the game. As a part of the introduction, participants are given three chances to work on a simple single-agent training task, to verify their understanding. Those who fail to complete the training task in one minute would not continue the study. This is a comprehension test to exclude people who do not understand game control.

Participants who finish training get to see further instructions before starting to collaborate with the robot. They are first educated about the goal of a collaboration task (i.e., making \textit{apple juice}) and what actions the team should perform to finish it. This is done to make sure every participant has sufficient knowledge to finish the task, so that the impact of user-specific prior knowledge can be minimized. To prepare users to interact and communicate with the robot agent, we would also show them a top-down view of the level map (as shown in Figure \ref{fig:level}), the appearance of the robot agent as well as an example of an explanation. During the task, the team is required to make and serve two orders of dishes in the virtual kitchen. At the end of the study, each participant is asked to complete a post-experiment survey to provide background information and evaluate the robot teammate.

\begin{figure}[t]
	\centering
	\includegraphics[width=0.9\linewidth]{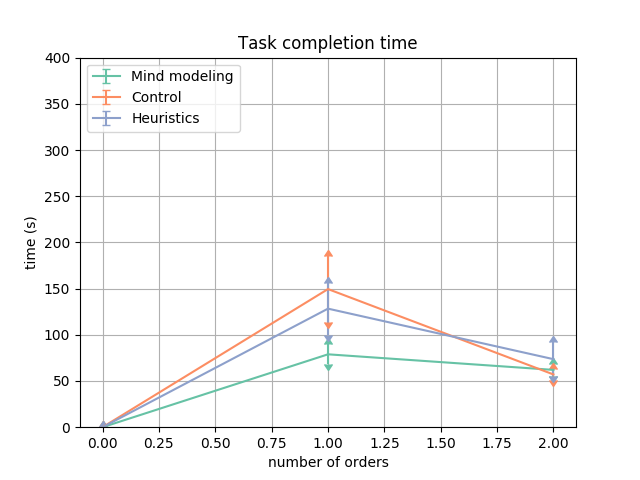}
	\caption{Time taken for the team to complete two orders under different testing conditions. }
	\label{fig:task_time}
	\vspace{-3pt}
\end{figure}

\noindent \textbf{Measurement.}
In the background study, we have collected from users their basic demographic information, education, as well as experience with video games. 

Our objective measure is intended to evaluate the human-robot teaming performance and subjective measure is designed for evaluating users' perception of the robot. Our dependent measures are listed below:
\begin{itemize}[leftmargin=*]
	\item \textbf{Teaming performance.} We evaluate teaming performance by recording the time for the team to complete each order. 
	\item \textbf{Perception of the robot.} We measure user's perception about the robot, in terms of its helpfulness and efficiency. Helpfulness is comprised of questions that measure users' opinion on the robot's ability to provide necessary help. Efficiency is comprised of questions that measure users' opinion on how efficiently and fluently the team is able to finish the task.
\end{itemize}

\begin{figure}[t]
	\begin{subfigure}[t]{0.24\textwidth}
		\includegraphics[width=\textwidth]{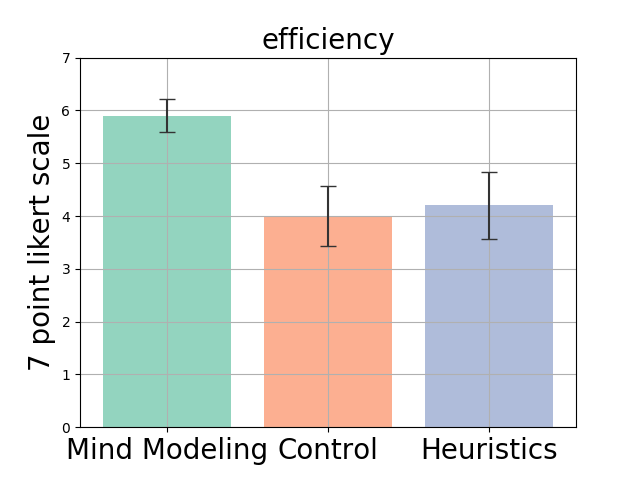}
	\end{subfigure}
	\begin{subfigure}[t]{0.24\textwidth}
		\includegraphics[width=\textwidth]{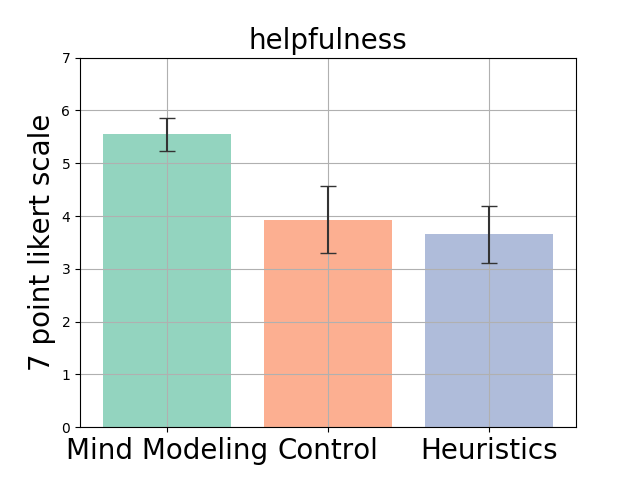}
	\end{subfigure}
	\caption{User's self-reported perception of the robot in terms of its efficiency and helpfulness.}
	\label{fig:percept}
	\vspace{-5pt}
\end{figure}

\subsection{Results and Analysis}

We recruited 29 subjects for our IRB-approved study from the university's subject pool. Most of the participants (69.3\%) came from a non-STEM background. Their reported ages ranged from 17 to 36 (M=19.52, SD=2.89). All the participants have moderate experience with video games and have not played the video game \textbf{Overcooked}, which inspired our study design.  Each participant got 1 course credit after completing the study. In addition, for ease of conducting the study, we discarded the data of 2 participants from the control group, as they got completely lost and failed to finish the designated task. As a result, there are 10 valid participants in the "mind modeling" and "heuristics" group, and 7 in the "control" group. 

Generally, we use ANOVA to test the effects of different experimental conditions on teaming performance and subjective perception of the robot. Tukey HSD tests are conducted on all possible pairs of experimental conditions.

As shown in Figure \ref{fig:task_time}, we found marginally significant effects from "mind modeling" conditions on completion time of the first order ($F(2, 24)=2.038, p=.152$). Post-hoc comparisons using the Tukey HSD tests revealed that teams could finish the first order significantly faster if users were under the "mind modeling" condition, compared to those under "control" ($p=.044$). The result is marginally significant compared to those in "heuristics" ($p=.120$), \textbf{confirming H1}. However, for the completion time of the second order, we did not find any significant effect ($F(2, 24)=0.425, p=.658$). This is not surprising since users were asked to finish the same task twice. They could take advantage of their previous experience working with the robot for the second order. Intuitively, the quantitative result showed that our explanation generation algorithm helped non-expert users to finish the task efficiently on their first run, while those in the control group needed to complete the task once to be able to finish it with the same efficiency.

The factorial ANOVA also revealed a significant effect of the explanation system on the perceived helpfulness ($F(2, 24)=4.663, p=.019$) and efficiency ($F(2, 24)=4.136, p=.029$) of the robot (Figure \ref{fig:percept}). \textbf{In support of H2}, post-hoc analysis with the Tukey HSD tests showed that the robot's perceived helpfulness was significantly higher under the "mind modeling" condition, compared to "control" ($p=.023$) and "heuristics" ($p<.01$). Users under the "mind modeling" were also more likely to believe the explanation system resulted in improved collaboration efficiency, compared to "heuristics" ($p=.026$) and "control" ($p<.01$). 

\section{Conclusion}
In this paper, we propose a framework that allows a robot agent to improve teaming performance by communicating compelling explanations to its non-expert human teammate. By maintaining the mental state of both agents, the robot agent successfully generates explanations when the human behavior deviates from the optimal plan. By conducting a user study on a virtual collaborative cooking task, we demonstrate that the proposed algorithm can improve efficiency and quality of the interaction. 

For simplicity of implementation, the current environment configuration prevents human and robot from having a shared workspace. For future work, we plan to study more cooking tasks in a diverse set of environments where multiple collaboration strategies can evolve. In addition, to make the robot's model more transparent, we consider to generate contrastive explanations with respect to identified incorrect user beliefs from the user's mental model in the future. Meanwhile, we plan to focus on a more balanced settings where both the human and robot agent have some information (e.g. ability, preference) to share with the teammates before a valid and efficient joint task plan can be formed.  

\section{Acknowledgements}
This work has been supported by DARPA XAI N66001-17-2-4029.

\bibliographystyle{IEEEtran}
\bibliography{IEEEexample}

\end{document}